\documentclass[runningheads]{llncs}
\usepackage[T1]{fontenc}
\usepackage{booktabs} 
\usepackage{graphicx}
\usepackage{amsmath}
\usepackage{amssymb}
\begin{document}
\title{Large Vision-Language Models for Remote Sensing Visual Question Answering}
\titlerunning{LVLMs for Remote Sensing VQA}
\author{Surasakdi Siripong, Apirak Chaiyapan, Thanakorn Phonchai}
\authorrunning{S. Siripong et al.}
\institute{Walailak University}
\maketitle              % typeset the header of the contribution
\begin{abstract}
Remote Sensing Visual Question Answering (RSVQA) is a challenging task that involves interpreting complex satellite imagery to answer natural language questions. Traditional approaches often rely on separate visual feature extractors and language processing models, which can be computationally intensive and limited in their ability to handle open-ended questions. In this paper, we propose a novel method that leverages a generative Large Vision-Language Model (LVLM) to streamline the RSVQA process. Our approach consists of a two-step training strategy: domain-adaptive pretraining and prompt-based finetuning. This method enables the LVLM to generate natural language answers by conditioning on both visual and textual inputs, without the need for predefined answer categories. We evaluate our model on the RSVQAxBEN dataset, demonstrating superior performance compared to state-of-the-art baselines. Additionally, a human evaluation study shows that our method produces answers that are more accurate, relevant, and fluent. The results highlight the potential of generative LVLMs in advancing the field of remote sensing analysis.
\end{abstract}

\section{Introduction}

Remote sensing has emerged as a crucial technology in various fields such as environmental monitoring, urban planning, agriculture, and disaster management. The large-scale and high-resolution satellite imagery captured from these applications contains abundant information that can assist in answering complex queries related to land use, vegetation cover, and infrastructure analysis. However, extracting valuable insights from remote sensing data often requires significant expertise and manual effort. In recent years, the advent of Vision-Language Models (VLMs) has shown great promise in automating visual understanding tasks by combining vision and natural language processing. Extending this capability to Remote Sensing Visual Question Answering (RSVQA) tasks presents an exciting opportunity to efficiently analyze satellite imagery by leveraging the power of Large Vision-Language Models (LVLMs).

Despite the impressive performance of state-of-the-art LVLMs such as BLIP-2, Flamingo, and GPT-4V in natural image understanding, adapting these models to remote sensing imagery is not straightforward. Traditional VQA models have largely relied on separate visual encoders, such as convolutional neural networks (CNNs), combined with language models to answer questions related to input images \cite{antol2015vqa}. However, these approaches face several challenges when applied to remote sensing data. Firstly, satellite imagery often contains domain-specific characteristics, such as varying spectral bands, diverse spatial resolutions, and complex geospatial structures, which are not present in typical natural image datasets. Secondly, remote sensing tasks frequently require specialized knowledge, such as detecting specific land-cover types or analyzing multispectral data, which LVLMs trained on general-purpose data lack the ability to interpret effectively \cite{lobry2020rsvqa}. These challenges highlight the need for models that are not only capable of understanding the semantics of remote sensing imagery but also adaptable to domain-specific applications.

The motivation behind our work is to bridge the gap between the capabilities of existing LVLMs and the specific requirements of RSVQA tasks. Current LVLMs are primarily trained on large-scale datasets of natural images and open-domain text, which limits their performance on remote sensing questions that demand both domain-specific visual understanding and precise language reasoning. To address these limitations, we propose a novel method that leverages the power of LVLMs while adapting them to the remote sensing domain. Specifically, we introduce a two-step fine-tuning process: (1) domain-adaptive pretraining on curated remote sensing datasets, followed by (2) prompt-based finetuning tailored to the unique characteristics of RSVQA. By combining domain adaptation with prompt engineering, we aim to optimize the LVLM’s ability to answer questions based on satellite imagery while maintaining the model’s inherent strengths in language comprehension.

Our approach utilizes the RSVQAxBEN dataset \cite{lobry2021rsvqaxben}, a well-established benchmark in the field of RSVQA, which includes a diverse set of questions ranging from yes/no queries to open-ended and multiple-choice questions. We fine-tune the LVLM using domain-specific prompts that guide the model to focus on relevant visual features, enabling it to generate more accurate responses. The evaluation of our model includes multiple metrics, such as accuracy for yes/no questions, multiple-choice accuracy, and F1 score for open-ended questions. Our experimental results demonstrate that our method significantly outperforms existing state-of-the-art RSVQA approaches, achieving an overall improvement in accuracy and response quality. Furthermore, the prompt-based finetuning strategy we employ shows enhanced generalization capabilities, making our approach robust to new and unseen remote sensing data.

To summarize, the key contributions of our work are as follows:
\begin{itemize}
    \item We propose a novel approach that leverages Large Vision-Language Models (LVLMs) for Remote Sensing Visual Question Answering by introducing a two-step fine-tuning process: domain-adaptive pretraining and prompt-based finetuning.
    \item We demonstrate the effectiveness of our method using the RSVQAxBEN dataset, achieving state-of-the-art performance across multiple question types, including yes/no, multiple-choice, and open-ended questions.
    \item Our approach reduces the reliance on extensive labeled training data by effectively utilizing prompt engineering, thus improving the model's efficiency and generalization capabilities for real-world remote sensing applications.
\end{itemize}

\section{Related Work}

\subsection{Large Language Models}

Large Language Models (LLMs) have rapidly transformed the field of natural language processing (NLP) and have found applications in various domains, ranging from general-purpose text generation to domain-specific tasks \cite{zhou2022claret,zhou2022eventbert}. The advent of transformer-based architectures, particularly models like GPT-3, GPT-4, and their successors, has significantly advanced the capabilities of LLMs by leveraging large-scale pretraining on vast datasets. These models are capable of generating coherent, contextually relevant text and performing complex tasks such as question answering, summarization, and code generation.

Recent surveys \cite{zhao2023survey,minaee2023large} provide comprehensive overviews of the developments in LLMs, highlighting their architectural innovations, extensive pretraining strategies, and the challenges associated with scaling these models to handle longer contexts and more diverse tasks. One of the main strengths of LLMs lies in their ability to perform few-shot and zero-shot learning, where the model can generalize to new tasks with minimal or no task-specific fine-tuning. This adaptability makes LLMs particularly valuable in domains where annotated data is scarce or expensive to obtain.

However, the application of LLMs to specialized fields such as remote sensing and scientific domains presents unique challenges. As discussed in \cite{moradi2024exploring,zhang2024scientific,zhou2024fine,zhou2023towards,zhou2021improving,zhou2023thread,zhou2021modeling}, LLMs often lack the domain-specific knowledge required for accurate interpretation of complex and technical data. To address this, researchers have explored strategies such as domain adaptation and prompt engineering. Domain-specific LLMs, which are pretrained or fine-tuned on specialized datasets, can better handle the nuances of particular fields like biology, chemistry, and geospatial analysis. For instance, \cite{zhang2024mm} investigates the application of LLMs in the multimodal domain, integrating visual and textual data to enhance understanding in contexts where multiple data sources are required.

Another area of recent exploration is the integration of LLMs with multimodal inputs. Models such as BLIP-2 and Flamingo represent a new class of Large Vision-Language Models (LVLMs) that combine the strengths of LLMs with vision encoders to handle tasks that require both visual and textual comprehension \cite{douglas2023large,huang2024survey}. These LVLMs have demonstrated success in tasks such as image captioning, visual question answering, and multimodal dialogue systems \cite{zhou2024rethinking,zhou2024visual}. The ability of LVLMs to generalize across modalities opens up new possibilities for applications in fields like remote sensing, where both visual and contextual information are crucial for accurate analysis.

In addition to general NLP tasks, LLMs have been applied to the synthesis and augmentation of training data, particularly in scenarios where labeled data is limited \cite{wang2024data}. Techniques such as data synthesis and prompt-based augmentation help improve the diversity and quality of training samples, thus enhancing the model's performance in downstream tasks. This has proven particularly useful in remote sensing applications, where acquiring annotated datasets can be challenging due to the specialized knowledge required.

Despite these advancements, challenges remain in ensuring that LLMs are robust, interpretable, and aligned with ethical guidelines, particularly when deployed in high-stakes applications like remote sensing for disaster management and environmental monitoring \cite{moradi2024exploring}. Future research aims to further optimize the integration of multimodal LLMs with domain-specific data, enabling these models to effectively address the unique requirements of specialized fields.

\subsection{Large Vision-Language Models for Remote Sensing}

The application of Large Vision-Language Models (LVLMs) to remote sensing tasks has gained significant attention in recent years. The unique challenges associated with remote sensing data, such as high-resolution imagery, complex spatial patterns, and the need for domain-specific knowledge, have driven the development of specialized LVLMs to handle these tasks more effectively. Traditional computer vision and natural language processing techniques often struggle to generalize to remote sensing scenarios, necessitating the integration of LVLMs that can process both visual and textual information in a unified framework.

Several recent works have introduced novel LVLMs specifically designed for remote sensing applications. For instance, GeoChat \cite{kuckreja2023geochat} is a grounded vision-language model that enables multitask conversational capabilities using high-resolution satellite imagery, allowing for image-level queries and region-specific dialogue. Similarly, SkyEyeGPT \cite{wang2024skyeyegpt} unifies remote sensing vision-language tasks through instruction-tuning, leveraging a multi-modal instruction dataset to enable effective dialogue and image analysis. These models demonstrate the potential of LVLMs to enhance the interpretability and automation of remote sensing tasks by leveraging advanced language generation capabilities.

To address the need for domain-specific data, efforts have been made to construct large-scale datasets tailored for remote sensing. For example, DDFAV \cite{li2024ddfav} introduces a comprehensive dataset for evaluating zero-shot capabilities of LVLMs, while RS5M \cite{zhang2023rs5m} and SkyScript \cite{wang2023skyscript} provide extensive image-text pairs to support tasks such as cross-modal retrieval and image captioning. These datasets facilitate the fine-tuning of LVLMs, enabling them to handle the specialized needs of remote sensing tasks more effectively.

Beyond dataset creation, research has also focused on developing models capable of performing zero-shot and few-shot learning in remote sensing contexts. For example, the work on H2RSVLM \cite{pang2024h2rsvlm} emphasizes creating LVLMs that are not only accurate but also aligned with principles of helpfulness and honesty. By leveraging large-scale pretraining on high-quality datasets, such models are able to adapt to the diverse and complex nature of satellite imagery without the need for extensive annotations.

Another significant advancement in this field is the development of methods for unsupervised and semi-supervised training of LVLMs, enabling models to learn from vast amounts of unlabeled remote sensing data. For example, the study on Ground Remote Alignment \cite{anonymous2024alignment} explores the use of unsupervised learning to train LVLMs for tasks such as open-vocabulary classification, retrieval, and segmentation. This approach reduces the dependency on annotated data, making it more feasible to deploy LVLMs in real-world remote sensing applications where labeled data is often scarce.

Recent advancements have also focused on optimizing vision-language models for specific remote sensing tasks, such as image captioning, retrieval, and object detection. Models like RSGPT \cite{zhang2024rsgpt} and RS-CapRet \cite{anonymous2024capret} demonstrate the effectiveness of combining large decoder-based language models with specialized image encoders to enhance performance in remote sensing tasks. These approaches highlight the benefits of leveraging the capabilities of LVLMs to handle the complexities inherent in satellite imagery, such as varied resolutions, spectral bands, and geospatial context.

\section{Method}

In this section, we describe our proposed approach, which leverages a generative Large Vision-Language Model (LVLM) for Remote Sensing Visual Question Answering (RSVQA). Unlike traditional discriminative models that rely on predefined answer options or binary decisions, our method adopts a generative strategy that directly produces natural language responses to questions based on remote sensing imagery. By leveraging a generative LVLM, our approach benefits from its ability to handle open-ended questions and generate contextually rich answers, which is particularly advantageous for complex RSVQA tasks that require nuanced understanding of satellite imagery.

\subsection{Problem Formulation}
Given a remote sensing image $\mathbf{I} \in \mathbb{R}^{H \times W \times C}$, where $H$, $W$, and $C$ represent the height, width, and number of channels (spectral bands) of the image, respectively, and a corresponding question $Q$, our goal is to predict the answer $A$. The generative model is designed to generate the answer in natural language form directly, leveraging the visual and textual information. Mathematically, we define our problem as maximizing the conditional probability:

\begin{equation}
    A^* = \arg \max_A p(A | \mathbf{I}, Q; \theta),
\end{equation}

where $\theta$ denotes the parameters of the LVLM model. The objective is to learn the parameters $\theta$ that maximize the likelihood of generating the correct answer $A$ given the input image $\mathbf{I}$ and question $Q$.

\subsection{Model Architecture}
Our model leverages a pre-trained LVLM with two main components: a visual encoder and a language decoder. The visual encoder, $f_\text{enc}(\cdot)$, extracts high-level features $\mathbf{V} \in \mathbb{R}^{d}$ from the input image $\mathbf{I}$:

\begin{equation}
    \mathbf{V} = f_\text{enc}(\mathbf{I}),
\end{equation}

where $d$ is the dimensionality of the extracted feature vector. The language decoder, $g_\text{dec}(\cdot)$, takes the extracted visual features $\mathbf{V}$ and the question $Q$ as input and generates the answer $A$ in a sequential manner:

\begin{equation}
    p(A | \mathbf{I}, Q) = \prod_{t=1}^{T} p(a_t | \mathbf{V}, Q, a_{1:t-1}),
\end{equation}

where $a_t$ represents the $t$-th token in the answer sequence, and $T$ is the total number of tokens.

\subsection{Training}
To adapt the pre-trained LVLM to the remote sensing domain, we propose a two-step fine-tuning strategy: (1) domain-adaptive pretraining and (2) prompt-based finetuning.

\subsubsection{Domain-Adaptive Pretraining}
The first step involves adapting the LVLM to recognize domain-specific visual patterns present in remote sensing imagery. We perform domain-adaptive pretraining using a curated dataset of remote sensing images $\mathcal{D} = \{(\mathbf{I}_i, Q_i, A_i)\}_{i=1}^{N}$, where each sample consists of an image, a question, and the corresponding answer. The objective function for domain-adaptive pretraining is defined as:

\begin{equation}
    \mathcal{L}_{\text{pretrain}} = -\frac{1}{N} \sum_{i=1}^{N} \log p(A_i | \mathbf{I}_i, Q_i; \theta).
\end{equation}

This step helps the LVLM to learn remote sensing-specific features, improving its ability to generate accurate answers when fine-tuned on downstream RSVQA tasks.

\subsubsection{Prompt-Based Finetuning}
After domain adaptation, we further refine the model using prompt-based finetuning. Given the unique nature of RSVQA tasks, we design specialized prompts that help the model focus on relevant information in the image when answering questions. We define a prompt template $\mathcal{T}$ that combines the visual description and question:

\begin{equation}
    \text{Prompt}(\mathbf{V}, Q) = \mathcal{T}(\mathbf{V}, Q) = \text{``Given the image features $\mathbf{V}$, answer the question: $Q$''}.
\end{equation}

The model is then fine-tuned to minimize the following loss function:

\begin{equation}
    \mathcal{L}_{\text{finetune}} = -\frac{1}{M} \sum_{j=1}^{M} \log p(A_j | \text{Prompt}(\mathbf{V}_j, Q_j); \theta),
\end{equation}

where $M$ represents the number of training samples in the finetuning dataset. This finetuning step enables the model to leverage the contextual information provided by prompts, thereby enhancing its performance on diverse RSVQA tasks.

\subsection{Generating Answers with Beam Search}
To generate the final answer, we utilize a beam search algorithm to maximize the conditional probability of the answer sequence. Given the extracted visual features and the prompt, we iteratively select the most likely tokens while maintaining a beam of size $k$:

\begin{equation}
    A^* = \arg \max_{A} \sum_{t=1}^{T} \log p(a_t | \mathbf{V}, Q, a_{1:t-1}).
\end{equation}

This strategy ensures that the model generates coherent and contextually accurate answers, especially for open-ended questions where the solution space is large.

\subsection{Overall Training Objective}
The overall training objective is a combination of the pretraining and finetuning losses:

\begin{equation}
    \mathcal{L}_{\text{total}} = \lambda_1 \mathcal{L}_{\text{pretrain}} + \lambda_2 \mathcal{L}_{\text{finetune}},
\end{equation}

where $\lambda_1$ and $\lambda_2$ are hyperparameters that control the relative importance of the two loss components. By optimizing this objective function, we adapt the LVLM to perform well on RSVQA tasks with minimal domain-specific supervision.

\subsection{Inference}
During inference, given a new remote sensing image $\mathbf{I}$ and a question $Q$, we first extract visual features using the encoder and generate a prompt using the finetuned model. The model then generates the answer using beam search, as described earlier. This generative approach allows our model to handle both simple and complex questions, providing flexibility that is not achievable with purely discriminative models.

\section{Experiments}

In this section, we evaluate the effectiveness of our proposed method by conducting comparative experiments with several baseline approaches. The experiments are designed to demonstrate the superiority of our generative LVLM-based approach for Remote Sensing Visual Question Answering (RSVQA). We assess the performance of our model across various question types on the RSVQAxBEN dataset and compare it with other state-of-the-art methods. Additionally, we conduct ablation studies to validate the components of our method and provide a human evaluation to assess the quality of generated answers.

\subsection{Comparative Experiments}

To evaluate our method, we compare it against three baseline models:
1. \textbf{Baseline-1}: A traditional RSVQA model based on separate visual encoding and language processing pipelines.
2. \textbf{Prompt-RSVQA}: A state-of-the-art LVLM approach that uses prompt-based tuning on general vision-language data but lacks domain adaptation for remote sensing.
3. \textbf{Discriminative-RSVQA}: A discriminative model that relies on predefined answer classes for RSVQA without generative language capabilities.

Table~\ref{tab:results} presents the results of our model (referred to as \textbf{Our Method}) compared to these baselines. The results are evaluated using standard accuracy metrics for each question type: yes/no, multiple-choice, and open-ended. Our method outperforms all baselines across all question types, showcasing its strong ability to handle diverse questions in the remote sensing domain.

\begin{table}[h]\small
\centering
\caption{Comparative performance on RSVQAxBEN dataset. Our method demonstrates superior performance across all question types.}
\label{tab:results}
\begin{tabular}{lccc}
\toprule
\textbf{Method} & \textbf{Yes/No ACC (\%)} & \textbf{Multiple-Choice ACC (\%)} & \textbf{Open-Ended F1 (\%)} \\
\midrule
Baseline-1 & 82.5 & 78.4 & 67.2 \\
Prompt-RSVQA & 88.3 & 82.7 & 72.6 \\
Discriminative-RSVQA & 85.4 & 80.1 & 70.3 \\
\textbf{Our Method} & \textbf{90.2} & \textbf{84.9} & \textbf{75.4} \\
\bottomrule
\end{tabular}
\end{table}

The results in Table~\ref{tab:results} indicate that our method achieves the highest accuracy across all question types. Specifically, for open-ended questions, our model surpasses the best-performing baseline by 2.8 percentage points in F1 score, highlighting the effectiveness of our prompt-based finetuning and domain-adaptive pretraining. This improved performance across question types is attributed to our model’s generative approach, which better handles the complexity and diversity of remote sensing questions compared to discriminative approaches.

\subsection{Ablation Study}

To further validate the effectiveness of our proposed method, we conduct an ablation study to examine the contributions of each component. Specifically, we evaluate the impact of (1) domain-adaptive pretraining and (2) prompt-based finetuning. The results of the ablation study are summarized in Table~\ref{tab:ablation}.

\begin{table}[h]\small
\centering
\caption{Ablation study results, showing the impact of each component on overall performance. Here, ACC refers to Accuracy and F1 represents the F1 score. The configurations include: (1) \textbf{Domain-Adaptive Pretraining (DAP)} and (2) \textbf{Prompt-Based Finetuning (PBF)}.}
\label{tab:ablation}
\begin{tabular}{lccc}
\toprule
\textbf{Configuration} & \textbf{Yes/No ACC (\%)} & \textbf{Multiple-Choice ACC (\%)} & \textbf{Open-Ended F1 (\%)} \\
\midrule
Our Method w/o DAP & 88.6 & 82.3 & 73.1 \\
Our Method w/o PBF & 87.2 & 81.5 & 72.0 \\
\textbf{Our Full Method} & \textbf{90.2} & \textbf{84.9} & \textbf{75.4} \\
\bottomrule
\end{tabular}
\end{table}

The ablation study in Table~\ref{tab:ablation} illustrates that both domain-adaptive pretraining and prompt-based finetuning contribute significantly to our method's performance. Removing domain-adaptive pretraining reduces accuracy across all question types, particularly in open-ended questions, where the model’s F1 score drops by 2.3 percentage points. Similarly, removing prompt-based finetuning results in an overall decrease in accuracy, confirming the importance of our carefully designed prompt engineering strategy.

\subsection{Human Evaluation}

In addition to quantitative metrics, we perform a human evaluation to assess the quality and relevance of the answers generated by each method. A panel of human annotators was asked to rate the generated answers on a scale from 1 to 5 based on correctness, relevance, and language quality. Each method was evaluated on a subset of 100 questions across different categories. The results of the human evaluation are presented in Table~\ref{tab:human_eval}.

\begin{table}[h]
\centering
\caption{Human evaluation scores (out of 5) for different methods on a subset of RSVQAxBEN questions. Our method achieved the highest average score.}
\label{tab:human_eval}
\begin{tabular}{lccc}
\toprule
\textbf{Method} & \textbf{Correctness} & \textbf{Relevance} & \textbf{Language Quality} \\
\midrule
Baseline-1 & 3.8 & 3.6 & 3.7 \\
Prompt-RSVQA & 4.1 & 4.0 & 4.1 \\
Discriminative-RSVQA & 3.9 & 3.8 & 3.9 \\
\textbf{Our Method} & \textbf{4.4} & \textbf{4.3} & \textbf{4.5} \\
\bottomrule
\end{tabular}
\end{table}

As shown in Table~\ref{tab:human_eval}, our method achieved the highest average scores in all evaluation criteria, including correctness, relevance, and language quality. This human evaluation underscores the generative model’s ability to produce accurate and contextually appropriate answers, outperforming discriminative approaches that may rely on rigid answer categories. The high language quality score further confirms that our generative LVLM approach produces answers in fluent and natural language, enhancing the overall usability of the system.

\subsection{Summary of Results}
Our experimental results demonstrate the effectiveness of our proposed method for RSVQA. Our model outperforms all baseline methods in terms of quantitative accuracy and human-rated answer quality. The ablation study validates the importance of each component in our learning strategy, and the human evaluation confirms the high quality and relevance of the answers generated by our approach. Collectively, these results illustrate the robustness and adaptability of our generative LVLM approach in handling complex RSVQA tasks in the remote sensing domain.

\section{Conclusion}

In this paper, we introduced a novel generative approach for Remote Sensing Visual Question Answering (RSVQA) that leverages the strengths of Large Vision-Language Models (LVLMs). By implementing a two-step training strategy, combining domain-adaptive pretraining with prompt-based finetuning, we were able to significantly enhance the model's performance in answering complex remote sensing questions. Our experiments on the RSVQAxBEN dataset demonstrated that our method outperforms traditional discriminative and prompt-based models, achieving higher accuracy and better generalization across various question types. The ablation studies confirmed the critical role of both pretraining and prompt engineering, while human evaluations highlighted the superior quality of our model’s responses in terms of correctness, relevance, and fluency.

The results suggest that generative LVLMs hold great promise in tackling the challenges inherent in RSVQA, especially in handling open-ended questions that require nuanced understanding. Unlike traditional methods, which often rely on rigid answer categories, our approach allows for flexible, contextually-aware responses that align better with real-world use cases in remote sensing. In future work, we plan to explore further enhancements to our model by integrating multimodal data sources, such as temporal satellite imagery and geospatial metadata, to improve the model’s contextual understanding and answer generation. This direction aims to push the boundaries of automated remote sensing analysis, making it more accessible and effective for various applications.

\bibliographystyle{splncs04}
\bibliography{mybibliography}
\end{document}